\documentclass[letterpaper]{article} 
\usepackage{aaai23}  
\usepackage{times}  
\usepackage{helvet}  
\usepackage{courier}  
\usepackage[hyphens]{url}  
\usepackage{graphicx} 
\usepackage{natbib}  
\usepackage{caption} 
\usepackage{algorithm}
\usepackage{algorithmic}
\usepackage{amssymb}
\usepackage{amsmath}
\usepackage{booktabs}
\usepackage{subcaption}
\usepackage{bbding}
\usepackage{multirow}
\usepackage[switch]{lineno}
\usepackage{amsthm}

\usepackage{newfloat}
\usepackage{listings}
\DeclareCaptionStyle{ruled}{labelfont=normalfont,labelsep=colon,strut=off} 
\floatstyle{ruled}
\newfloat{listing}{tb}{lst}{}
\floatname{listing}{Listing}
\title{OMPQ: Orthogonal Mixed Precision Quantization}
\author{
    Yuexiao Ma\textsuperscript{\rm 1},
    Taisong Jin\textsuperscript{\rm 2}\thanks{Corresponding Author: jintaisong@xmu.edu.cn},
    Xiawu Zheng\textsuperscript{\rm 3},
    Yan Wang\textsuperscript{\rm 4},
    Huixia Li\textsuperscript{\rm 2}, \\
    Yongjian Wu\textsuperscript{\rm 5},
    Guannan Jiang\textsuperscript{\rm 6},
    Wei Zhang\textsuperscript{\rm 6},
    Rongrong Ji\textsuperscript{\rm 1}
}
\affiliations{
    \textsuperscript{\rm 1} Media Analytics and Computing Lab, Department of Artificial Intelligence, School of Informatics, \\ Xiamen University, China.
    \textsuperscript{\rm 2} Media Analytics and Computing Lab, Department of Computer Science and Technology, \\ School of Informatics, Xiamen University, China.
    \textsuperscript{\rm 3} Peng Cheng Laboratory, Shenzhen, China. \\
    \textsuperscript{\rm 4} Samsara, Seattle, WA, USA.
    \textsuperscript{\rm 5} Tencent Youtu Lab, Shanghai, China.
    \textsuperscript{\rm 6} CATL, China.
\\
    bobma@stu.xmu.edu.cn, jintaisong@xmu.edu.cn, zhengxw01@pcl.ac.cn, grapeot@outlook.com, \\ hxlee@stu.xmu.edu.cn, littlekenwu@tencent.com, \{jianggn, zhangwei\}@catl.com, rrji@xmu.edu.cn
}

\usepackage{bibentry}

\begin{document}

\maketitle

\begin{abstract}
To bridge the ever-increasing gap between deep neural networks' complexity and hardware capability, network quantization has attracted more and more research attention. The latest trend of mixed precision quantization takes advantage of hardware's multiple bit-width arithmetic operations to unleash the full potential of network quantization. However, existing approaches rely heavily on an extremely time-consuming search process and various relaxations when seeking the optimal bit configuration. To address this issue, we propose to optimize a proxy metric of network orthogonality that can be efficiently solved with linear programming, which proves to be highly correlated with quantized model accuracy and bit-width. Our approach significantly reduces the search time and the required data amount by orders of magnitude, but without a compromise on quantization accuracy. Specifically, we achieve $72.08\%$ Top-1 accuracy on ResNet-$18$ with $6.7$Mb parameters, which does not require any searching iterations. Given the high efficiency and low data dependency of our algorithm, we use it for the post-training quantization, which achieves $71.27\%$ Top-1 accuracy on MobileNetV$2$ with only $1.5$Mb parameters.
\end{abstract}

\section{Introduction}
\label{sec:intro}

Recently, we witness an obvious trend in deep learning that the models have rapidly increasing complexity \cite{he2016deep,simonyan2014very,szegedy2015going,howard2017mobilenets,sandler2018mobilenetv2,zhang2018shufflenet}.
Due to practical limits such as latency, battery, and temperature, the host hardware where the models are deployed cannot keep up with this trend.
It results in a large and ever-increasing gap between the computational demands and the resources.
To address this issue, network quantization~\cite{courbariaux2016binarized,rastegari2016xnor,kim2019binaryduo,banner2019post,liu2019circulant}, which maps single-precision floating point weights or activations to lower bits integers for compression and acceleration, has attracted considerable research attention.
Network quantization can be naturally formulated as an optimization problem and a straightforward approach is to relax the constraints to make it a tractable optimization problem, at a cost of an approximated solution, 
\begin{figure} [tb]
\centering
\includegraphics[width=0.84\linewidth]{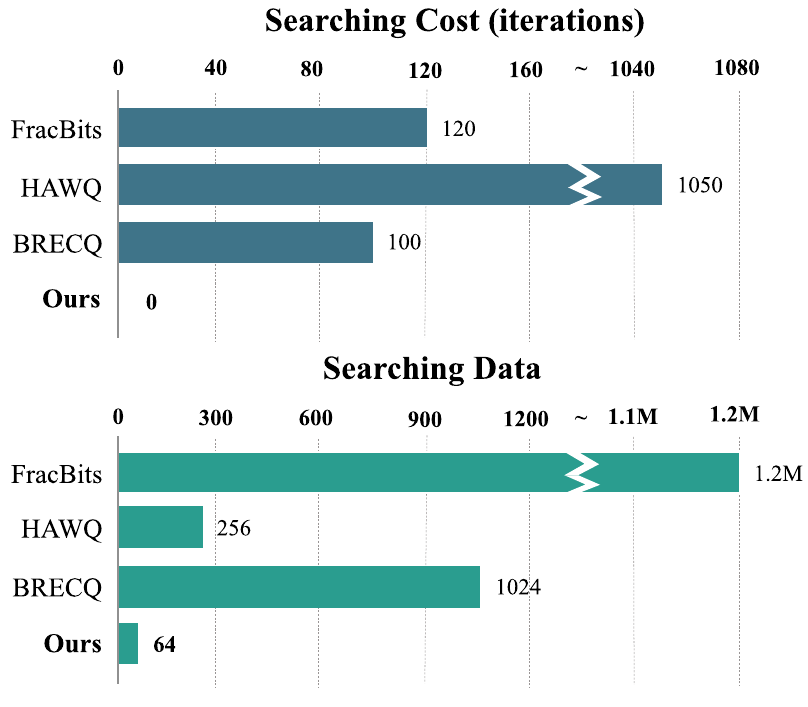}
\caption{Comparison of the resources used to obtain the optimal bit configuration between our algorithm and other mixed precision algorithms (FracBits \cite{yang2020fracbits}, HAWQ \cite{dong2019hawq}, BRECQ \cite{li2021brecq}) on ResNet-$18$. ``Searching Data" is the number of input images.}
\label{fig:data and cost}
\end{figure}
\emph{e.g.}~Straight Through Estimation (STE) \cite{bengio2013estimating}.

With the recent development of inference hardware, arithmetic operations with variable bit-width become a possibility and bring further flexibility to the network quantization.
To take full advantage of the hardware capability, mixed precision quantization \cite{dong2019hawq,wang2019haq,li2021brecq,yang2020fracbits} aims to quantize different network layers to different bit-width, so as to achieve a better trade-off between compression ratio and accuracy. 
While benefiting from the extra flexibility, the mixed precision quantization also suffers from a more complicated and challenging optimization problem, with a non-differentiable and extremely non-convex objective function.
Therefore, existing approaches~\cite{dong2019hawq,yang2020fracbits,wang2019haq,li2021brecq} often require numerous data and computing resources to search for the optimal bit configuration.
For instance, FracBits~\cite{yang2020fracbits} approximates the bit-width by performing a first-order Taylor expansion at the adjacent integer, making the bit variable differentiable. 
This allows it to integrate the search process into training to obtain the optimal bit configuration.
However, to derive a decent solution, it still requires a large amount of computation resources in the searching and training process.
To resolve the large demand on training data, Dong \emph{et al.}~\cite{dong2019hawq} use the average eigenvalue of the hessian matrix of each layer as the metric for bit allocation.
However, the matrix-free Hutchinson algorithm for implicitly calculating the average of the eigenvalues of the hessian matrix still needs 50 iterations for each network layer.
Another direction is black box optimization.
For instance, Wang \emph{et al.}~\cite{wang2019haq} use reinforcement learning for the bit allocation of each layer.
Li \emph{et al.}~\cite{li2021brecq} use evolutionary search algorithm \cite{guo2020single} to derive the optimal bit configuration, together with a block reconstruction strategy to efficiently optimize the quantized model. 
But the population evolution process requires $1,024$ input data and $100$ iterations, which is time-consuming.

Different from the existing approaches of black box optimization or constraint relaxation, we propose to construct a proxy metric, which could have a substantially different form, but be highly correlated with the objective function of original linear programming.
In general, we propose to obtain the optimal bit configuration by using the orthogonality of neural network. 
Specifically, we deconstruct the neural network into a set of functions, and define the orthogonality of the model by extending its definition from a function $ f:\mathbb{R}\rightarrow\mathbb{R} $ to the entire network $ f:\mathbb{R}^m\rightarrow\mathbb{R}^n $.
The measurement of the orthogonality could be efficiently performed with Monte Carlo sampling and Cauchy-Schwarz inequality, based on which we propose an efficient metric named \textbf{OR}thogonality \textbf{M}etric (ORM) as the proxy metric.
As illustrated in Fig.~\ref{fig:data and cost}, we only need a single-pass search process on a small amount of data with ORM. 
In addition, we derive an equivalent form of ORM to accelerate the computation.

On the other hand, model orthogonality and quantization accuracy are positively correlated on different networks. Therefore, maximizing model orthogonality is taken as our objective function. Meanwhile, our experiments show that layer orthogonality and bit-width are also positively correlated.
We assign a larger bit-width to the layer with larger orthogonality
while combining specific constraints to construct a linear programming problem. The optimal bit configuration can be gained simply by solving the linear programming problem.

In summary, our contributions are listed as follows:
\begin{itemize}
\item We introduce a novel metric of layer orthogonality. We introduce function orthogonality into neural networks and propose the \textbf{OR}thogonality \textbf{M}etric (ORM). We leverage it as a proxy metric to efficiently solve the mixed precision quantization problem, which is the first attempt in the community and can easily be integrated into any quantization scheme.
\item We observe a positive correlation between ORM and quantization accuracy on different models. Therefore, we optimize the model orthogonality through the linear programming problem, which can derive the optimal bit-width configuration in a few seconds without iterations. 
\item We also provide extensive experimental results on ImageNet, which elaborate that the proposed orthogonality-based approach could gain the state-of-the-art quantization performance with orders of magnitude's speed up.

\end{itemize}

\begin{figure*}[t]
	\centering
	\includegraphics[width=1\linewidth]{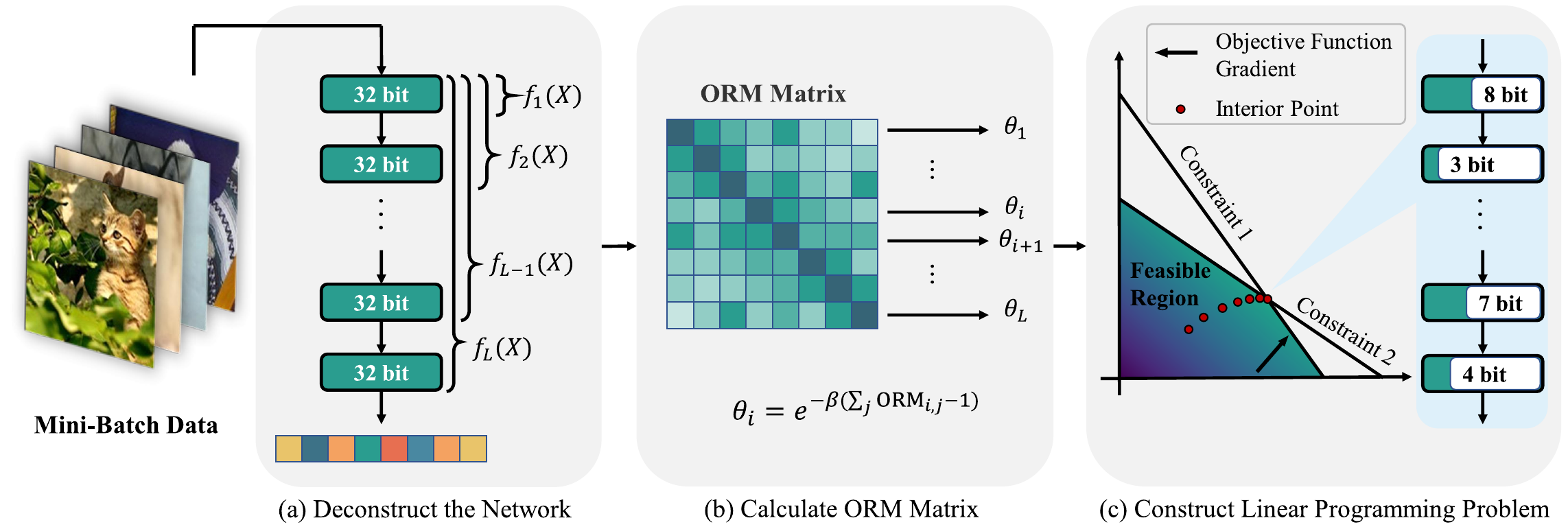}
	\caption{Overview. Left: Deconstruct the model into a set of functions $ \mathcal{F} $. Middle: ORM symmetric matrix calculated from $\mathcal{F}$. Right: Linear programming problem constructed by the importance factor $\theta$ to derive optimal bit configuration.}
	\label{fig:overview}
\end{figure*}

\section{Related Work}
\label{sec:formatting}

\noindent \textbf{Quantized Neural Networks:}
Existing neural network quantization algorithms can be divided into two categories based on their training strategy: post-training quantization (PTQ) and quantization-aware training (QAT). PTQ \cite{li2021brecq,cai2020zeroq,nagel2019data} is an offline quantization method, which only needs a small amount of data to complete the quantization process.
Therefore, PTQ could obtain an optimal quantized model efficiently, at a cost of accuracy drop from quantization.
In contrast, quantization-aware training \cite{gong2019differentiable,zhou2016dorefa,dong2019hawq,zhou2017incremental,chen2019metaquant,cai2017deep,choi2018pact} adopts online quantization strategy. This type of methods utilize the whole training dataset during quantization process.
As a result, it has superior accuracy but limited efficiency.

If viewed from a perspective of bit-width allocation strategy, neural network quantization can also be divided into unified quantization and mixed precision quantization.
Traditionally, network quantization means unified quantization.
Choi \emph{et al.}~\cite{choi2018pact} aim to optimize the parameterized clip boundary of activation value of each layer during training process.
Chen \emph{et al.}~\cite{chen2019metaquant} introduce the meta-quantization block to approximate the derivative of non-derivable function.
Recently, some works \cite{yang2020fracbits,dong2019hawq,li2021brecq} that explore assigning different bit-widths to different layers begin to emerge.
Yang \emph{et al.}~\cite{yang2020fracbits} approximate the derivative of bit-width by first-order Taylor expansion at adjacent integer points, thereby fusing the optimal bit-width selection with the training process. However, its optimal bit-width searching takes $80\%$ of the training epochs, which consumes lots of time and computation power.
Dong \emph{et al.}~\cite{dong2019hawq} take the average eigenvalues of the hessian matrix of each layer as the basis for the bit allocation of that layer.
However, the matrix-free Hutchinson algorithm for calculating the average eigenvalues needs to be iterated $50$ times for each layer.

\noindent \textbf{Network Similarity:} 
Previous works \cite{bach2002kernel,gretton2003kernel,leurgans1993canonical,fukumizu2004dimensionality,gretton2005measuring,kornblith2019similarity} define covariance and cross-covariance operators in the Reproducing Kernel Hilbert Spaces (RKHSs), and derive mutual information criteria based on these operators.
Gretton \emph{et al.}~\cite{gretton2005measuring} propose the Hilbert-Schmidt Independence Criterion (HSIC), and give a finite-dimensional approximation of it.
Furthermore, Kornblith \emph{et al.}~\cite{kornblith2019similarity} give the similarity criterion CKA based on HSIC, and study its relationship with the other similarity criteria.
In the following, we propose a metric from the perspective of network orthogonality, and give a simple and clear derivation. Simultaneously, we use it to guide the network quantization.

\section{Methodology}

In this section, we will introduce our mixed precision quantization algorithm from three aspects: how to define the orthogonality, how to efficiently measure it, and how to construct a linear programming model to derive the optimal bit configuration.

\subsection{Network Orthogonality}
\label{sec: Network Orthogonality}
A neural network can be naturally decomposed into a set of layers or functions. Formally, for the given input $x\in \mathbb{R}^{1\times (C\times H\times W)}$, we decompose a neural network into $ \mathcal{F}=\{f_1, f_2,\cdots, f_L\} $, where $ f_i $ represents the transformation from input $x$ \emph{to} the result of the $i$-th layer.
In other words, if $g_i$ represents the function \emph{of} the $i$-th layer, then $ f_i(x) = g_i\big(f_{i-1}(x)\big)=g_{i}\Big(g_{i-1}\big(\cdots g_1(x) \big) \Big)$.
Here we introduce the inner product \cite{arfken1999mathematical} between functions $ f_i $ and $f_j$, which is formally defined as,

\begin{equation}\label{eq:definition}
	{\left\langle f_i,f_j \right\rangle}_{P(x)} = \int_{\mathcal{D}} f_i(x)P(x){f_j(x)}^T dx, 
\end{equation}
where $ f_i(x)\in\mathbb{R}^{1\times(C_i\times H_i\times W_i)} $, $f_j(x)\in\mathbb{R}^{1\times(C_j\times H_j\times W_j)}$ are the known functions when the model is given, and $\mathcal{D}$ is the domain of $x$. If we set $f_i^{(m)}(x)$ to be the $m$-th element of $f_i(x)$, then $P(x)\in\mathbb{R}^{(C_i\times H_i\times W_i)\times(C_j\times H_j\times W_j)} $ is the probability density matrix between $f_i(x)$ and $f_j(x)$, where $P_{m,n}(x)$ is the probability density function of the random variable $ f_i^{(m)}(x)\cdot f_j^{(n)}(x) $.
According to the definition in \cite{arfken1999mathematical}, $ {\left\langle f_i,f_j \right\rangle}_{P(x)}=0 $ means that $ f_i $ and $ f_j $ are weighted orthogonal.
In other words, $ {\left\langle f_i,f_j \right\rangle}_{P(x)} $ is negatively correlated with the orthogonality between $ f_i $ and $ f_j $.
 When we have a known set of functions to be quantized $\mathcal{F} = \{ f_i \}_{i=1}^L$, with the goal to approximate an arbitrary function $h^*$,
the quantization error can then be expressed by the mean square error: $ \xi\int_{\mathcal{D}} {|h^*(x)-\sum_i \psi_if_i(x)|}^2dx $, where $\xi$ and $\psi_i$ are combination coefficient.
According to Parseval equality \cite{tanton2005encyclopedia}, if $\mathcal{F}$ is an orthogonal basis functions set, then the mean square error could achieve $0$.
Furthermore, the orthogonality between the basis functions is stronger, the mean square error is smaller, \emph{i.e.}, the model corresponding to the linear combination of basis functions has a stronger representation capability.
Here we further introduce this insight to network quantization.
In general, the larger the bit, the more representational capability of the corresponding model \cite{liu2018bi}.
Specifically, we propose to assign a larger bit-width to the layer with stronger orthogonality against all other layers to maximize the representation capability of the model.
However, Eq.~\ref{eq:definition} has the integral of a continuous function which is untractable in practice.
Therefore, we derive a novel metric to efficiently approximate the orthogonality of each layer in Sec~\ref{Efficient Orthogonality Metric}.

\subsection{Efficient Orthogonality Metric}
\label{Efficient Orthogonality Metric}

To avoid the intractable integral, we propose to leverage the Monte Carlo sampling to approximate the orthogonality of the layers. Specifically, from the Monte Carlo integration perspective in \cite{caflisch1998monte}, Eq.~\ref{eq:definition} can be rewritten as
\begin{equation}
\begin{split}
	{\left\langle f_i,f_j \right\rangle}_{P(x)}
	&= \int_{\mathcal{D}} f_i(x)P(x){f_j(x)}^T dx \\
	&= {\left\Vert E_{P(x)}[{f_j(x)}^Tf_i(x)]\right\Vert}_F \label{eq:montecarlo}.
\end{split}
\end{equation}
We randomly get $N$ samples $x_1,x_2,\dots,x_N $ from a training dataset with the probability density matrix $P(x)$, which allows the expectation $E_{P(x)}[{f_j(x)}^Tf_i(x)]$ to be further approximated as,
\begin{equation}
\begin{split}
	{\left\Vert E_{P(x)}[{f_j(x)}^Tf_i(x)] \right\Vert}_F&\approx \frac{1}{N} {\left\Vert\sum_{n=1}^{N} {f_j(x_n)}^Tf_i(x_n)\right\Vert}_F \\
	&= \frac{1}{N} {\left\Vert f_j(X)^{T}f_i(X)\right\Vert}_F \label{eq:discretization}, 
\end{split}
\end{equation}
where $ f_i(X) \in \mathbb{R}^{N \times(C_i\times H_i\times W_i)} $ represents the output of the $i$-th layer, $ f_j(X) \in \mathbb{R}^{N\times(C_j\times H_j\times W_j)} $ represents the output of the $j$-th layer, and $||\cdot||_F$ is the Frobenius norm. 
From Eqs.~\ref{eq:montecarlo}-\ref{eq:discretization}, we have

\begin{equation}\label{eq:Intermediate}
	N\int_{\mathcal{D}} f_i(x)P(x){f_j(x)}^T dx \approx {\left\Vert f_j(X)^{T}f_i(X)\right\Vert}_F. 
\end{equation}

\noindent However, the comparison of orthogonality between different layers is difficult due to the differences in dimensionality. To this end, we use the Cauchy-Schwarz inequality to normalize it in $[0,1]$ for the different layers. Applying Cauchy-Schwarz inequality to the left side of Eq.~\ref{eq:Intermediate}, we have

\begin{equation}
\begin{split}
\label{eq:Intermediate1}
	0 &\leq {\left( N\int_{\mathcal{D}} f_i(x)P(x){f_j(x)}^T dx \right)}^2 \\
	& \leq \int_{\mathcal{D}} Nf_i(x)P_i(x){f_i(x)}^T dx \int_{\mathcal{D}} Nf_j(x)P_j(x){f_j(x)}^T dx. 
\end{split}
\end{equation}
We substitute Eq.~\ref{eq:Intermediate} into Eq.~\ref{eq:Intermediate1} and perform some simplifications to derive our \textbf{OR}thogonality \textbf{M}etric (ORM)
\footnote{
ORM is formally consistent with CKA. However, we pioneer to discover its relationship with quantized model accuracy and confirm its validity in mixed precision quantization from the perspective of function orthogonality, and CKA explores the relationship between hidden layers from the perspective of similarity. In other words, CKA implicitly verifies the validity of ORM further.
}, refer to \textbf{supplementary material} for details:

\begin{equation}\label{eq:ORM}
	{\rm ORM}(X,f_i,f_j) = \frac{{||f_j(X)^{T}f_i(X)||}^2_F}{||f_i(X)^{T}f_i(X)||_F||f_j(X)^{T}f_j(X)||_F}, 
\end{equation}
where ORM $ \in [0,1] $. $f_i$ and $f_j$ is orthogonal when ${\rm \text{ORM}}=0$. On the contrary, $f_i$ and $f_j$ is dependent when ${\rm \text{ORM}}=1$. Therefore, ORM is negatively correlated to orthogonality. 

\textbf{Calculation Acceleration.} Given a specific model, calculating Eq.~\ref{eq:ORM} involves the huge matrices. Suppose that $f_i(X) \in \mathbb{R}^{N \times(C_i\times H_i\times W_i)} $, $ f_j(X) \in \mathbb{R}^{N\times (C_j\times H_j\times W_j)} $, and the dimension of features in the $j$-th layer is larger than that of the $i$-th layer. Furthermore, the time complexity of computing ${\rm ORM}(X,f_i,f_j)$ is $ \boldsymbol{O}(NC_j^2H_j^2W_j^2) $. The huge matrix occupies a lot of memory resources, and also increases the time complexity of the entire algorithm by several orders of magnitude. Therefore, we derive an equivalent form to accelerate calculation. If we take $ Y = f_i(X)$, $Z = f_j(X) $ as an example, then $ YY^T,ZZ^T \in \mathbb{R}^{N\times N} $. We have:

\begin{equation}\label{eq:efficiency_equal}
	||Z^TY||_F^2 = \left\langle \textbf{vec}(YY^T),\textbf{vec}(ZZ^T) \right\rangle, 
\end{equation}
where \textbf{vec($\cdot$)} represents the operation of flattening matrix into vector. From Eq.~\ref{eq:efficiency_equal}, 
the time complexity of calculating ${\rm ORM}(X,f_i,f_j)$ becomes $ \boldsymbol{O}(N^2C_jH_jW_j) $ through the inner product of vectors. When the number of samples $N$ is larger than the dimension of features $C\times H\times W$, the norm form is faster to calculate thanking to lower time complexity, vice versa. Specific acceleration ratio and the proof of Eq.~\ref{eq:efficiency_equal} are demonstrated in \textbf{supplementary material}.

\subsection{Mixed Precision Quantization}

\begin{figure}[tb]
\centering
\includegraphics[width=1.0\linewidth]{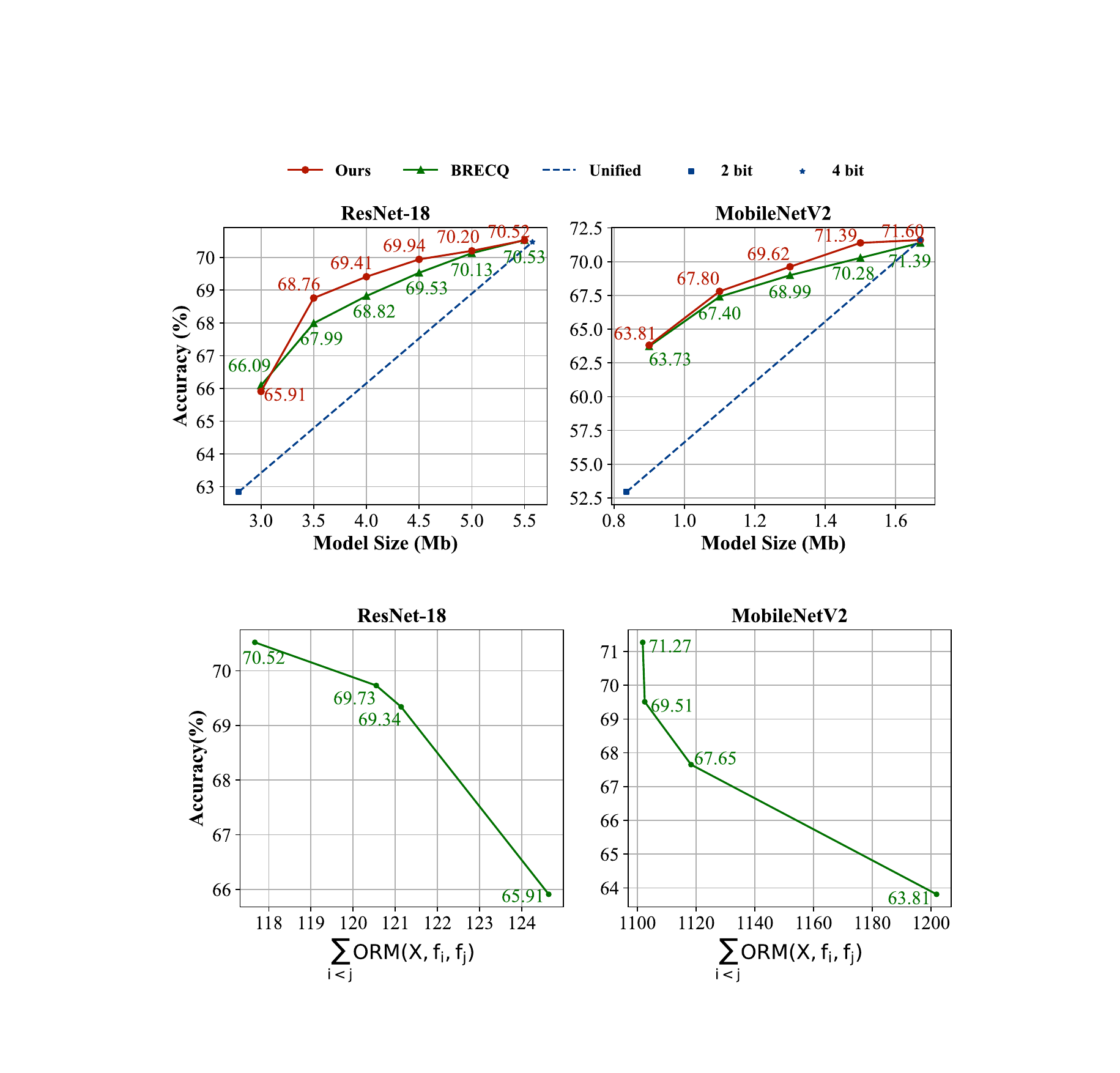}
\caption{Relationship between orthogonality and accuracy for different quantization configurations on ResNet-$18$ and MobileNetV$2$.}
\label{fig:orthogonality_accuracy}
\end{figure}

\noindent\textbf{Effectiveness of ORM on Mixed Precision Quantization.} ORM directly indicates the importance of the layer in the network, which can be used to decide the configuration of the bit-width eventually. We conduct extensive experiments to provide sufficient and reliable evidence for such claim. Specifically, we first sample different quantization configurations for ResNet-18 and MobileNetV2. Then finetuning to obtain the performance. Meanwhile, the overall orthogonality of the sampled models is calculated separately. Interestingly, we find that model orthogonality and performance are positively correlated to the sum of ORM in Fig.~\ref{fig:orthogonality_accuracy}. 
Naturally, inspired by this finding, maximizing orthogonality is taken as our objective function, which is employed to integrate the model size constraints and construct a linear programming problem to obtain the final bit configuration. 
The detailed experiments are provided in the \textbf{supplementary material}.

For a specific neural network, we can calculate an orthogonality matrix $K$, where $ k_{ij}={\rm ORM}(X, f_i, f_j) $. Obviously, $K$ is a symmetric matrix and the diagonal elements are $1$. Furthermore, we show some ORM matrices on widely used models with the different number of samples $N$ in the \textbf{supplementary material}. We add up the non-diagonal elements of each row of the matrix,

\begin{equation}\label{eq:layer_ort}
	\gamma_i = \sum_{j=1}^{L}k_{ij} - 1.
\end{equation}
Smaller $ \gamma_i $ means stronger orthogonality between $ f_i $ and other functions in the function set $ \mathcal{F} $, and it also means that former $i$ layers of the neural network are more independent. Thus, we leverage the monotonically decreasing function $ e^{-x} $ to model this relationship:

\begin{equation}\label{eq:layer_imp}
	\theta_i = e^{-\beta\gamma_i}, 
\end{equation}
where $ \beta $ is a hyper-parameter to control the bit-width difference between different layers. We also investigate the other monotonically decreasing functions (For the details, please refer to Sec~\ref{Ablation Study}). $ \theta_i $ is used as the importance factor for the former $i$ layers of the network, then we define a linear programming problem as follows:

\begin{equation}
\begin{split}
\label{linear_pro}
	\text{Objective:    }
	     &\max_{\mathbf{b}} \sum_{i=1}^{L}\left( \frac{b_i}{L-i+1} \sum_{j=i}^{L} \theta_j \right), \\
	\text{Constraints:    }
	     &  \sum_i^L M^{(b_i)} \leq \mathcal{T}.
\end{split}
\end{equation}
$ M^{(b_i)} $ is the model size of the $i$-th layer under $ b_i $ bit quantization and $\mathcal{T}$ represents the target model size. \textbf{b} is the optimal bit configuration. Maximizing the objective function means assigning the larger bit-width to more independent layer, which implicitly maximizes the model's representation capability. More details of network deconstruction, linear programming construction and the impact of $\beta$ are provided in the \textbf{supplementary material}. 

Note that it is extremely efficient to solve the linear programming problem in Eq.~\ref{linear_pro}, which only takes a few seconds on a single CPU. In other words, our method is extremely efficient ($9$s on MobileNetV$2$) when comparing to the previous methods \cite{yang2020fracbits,dong2019hawq,li2021brecq} that require lots of data or iterations for searching. In addition, our algorithm can be combined as a plug-and-play module with quantization-aware training or post-training quantization schemes thanking to the high efficiency and low data requirements. In other words, our approach is capable of improving the accuracy of SOTA methods, where detail results are reported in Secs~\ref{Quantization-Aware Training} and \ref{Post-Training Quantization}.

\begin{table}[!t]
\centering
\setlength{\tabcolsep}{2mm}{
\begin{tabular}{cccc}
\toprule  
Decreasing& 
ResNet-$18$ & 
MobileNetV$2$ & 
Changing \\
Function& 
($\%$)& 
($\%$)& 
Rate \\
\midrule  
$ e^{-x} $ & $72.30$ & $\textbf{63.51}$ &$ e^{-x} $\\
$ -logx $  & $72.26$ & $63.20$ &$ x^{-2} $\\
$ -x $     & $\textbf{72.36}$ & $63.0$  & $0$\\
$ -x^3 $   & $71.71$ & -     & $ 6x $\\
$ -e^x $   & -     & -     & $ e^x $\\
\bottomrule 
\end{tabular}}
\caption{The Top-1 accuracy (\%) with different monotonically decreasing functions on ResNet-$18$ and MobileNetV$2$.}
\label{tab:Decreasing Function}
\end{table}

\begin{table}[!t]
\centering
\setlength{\tabcolsep}{1.8mm}{
\begin{tabular}{cccccc}
\toprule  
Model & 
W bit& 
Layer&
Block &
Stage&
Net\\
\midrule  
ResNet-$18$   & $5*$ & $72.51$ & $\textbf{72.52}$ & $72.47$ & $72.31$ \\
MobileNetV$2$ & $3*$ & $\textbf{69.37}$  
& $69.10$ & $68.86$ & $63.99$ \\
\bottomrule 
\end{tabular}}
\caption{Top-$1$ accuracy (\%) of different deconstruction granularity. The activations bit-width of MobileNetV$2$ and ResNet-$18$ are both $8$. $*$ means mixed bit.}
\label{tab:granularity}
\end{table}

\begin{table*}[!t]
\centering

\setlength{\tabcolsep}{4mm}{
\begin{subtable}[ResNet-$18$]{1\linewidth}
\caption{ResNet-$18$}
\centering
\begin{tabular}{cr@{/}lccccc}
\toprule  
Method & 
W &
A &
Int-Only& 
Uniform& 
Model Size (Mb) &
BOPs (G)&
Top-1 ($\%$)\\
\midrule  
Baseline & $32$&$32$ & \XSolidBrush & - & $44.6$ & $1,858$ & $73.09$\\
\midrule  
RVQuant & $8$&$8$ & \XSolidBrush & \XSolidBrush & $11.1$ & $116$ & $70.01$\\
HAWQ-V3 & $8$&$8$ & \CheckmarkBold & \CheckmarkBold & $11.1$ & $116$ & $71.56$\\
OMPQ & $*$&$8$ & \CheckmarkBold & \CheckmarkBold & $6.7$ & $97$ & $\textbf{72.30}$\\
\midrule  
$\text{PACT}^\triangledown$ & $5$&$5$ & \XSolidBrush & \CheckmarkBold & $7.2$ & $74$ & $69.80$\\
$\text{LQ-Nets}^\triangledown$ & $4$&$32$ & \XSolidBrush & \XSolidBrush & $5.8$ & $225$ & $70.00$\\
HAWQ-V3 & $*$&$*$ & \CheckmarkBold & \CheckmarkBold & $6.7$ & $72$ & $70.22$\\
OMPQ & $*$&$6$ & \CheckmarkBold & \CheckmarkBold & $6.7$ & $75$ & $\textbf{72.08}$\\
\bottomrule 
\end{tabular}
\end{subtable}

\begin{subtable}[ResNet-$50$]{1\linewidth}
\caption{ResNet-$50$}
\centering
\begin{tabular}{cr@{/}lccccc}
\toprule  
Method & 
W &
A &
Int-Only& 
Uniform& 
Model Size (Mb) &
BOPs (G)&
Top-1 ($\%$)\\
\midrule  
Baseline & $32$ & $32$ & \XSolidBrush & - & $97.8$ & $3,951$ & $77.72$\\
\midrule  
$\text{PACT}^\triangledown$ & $5$ & $5$ & \XSolidBrush & \CheckmarkBold & $16.0$ & $133$ & $\textbf{76.70}$\\
$\text{LQ-Nets}^\triangledown$ & $4$ & $32$ & \XSolidBrush & \XSolidBrush & $13.1$ & $486$ & $76.40$\\
RVQuant & $5$ & $5$ & \XSolidBrush & \XSolidBrush & $16.0$ & $101$ & $75.60$\\
HAQ & * & $32$ & \XSolidBrush & \XSolidBrush & $9.62$ & $520$ & $75.48$\\
OneBitwidth & * & $8$ & \XSolidBrush & \CheckmarkBold & $12.3$ & $494$ & $\textbf{76.70}$\\
HAWQ-V3 & * & * & \CheckmarkBold & \CheckmarkBold & $18.7$ & $154$ & $75.39$\\
OMPQ & * & $5$ & \CheckmarkBold & \CheckmarkBold & $16.0$ & $141$ & $76.20$\\
OMPQ & * & $5$ & \CheckmarkBold & \CheckmarkBold & $18.7$ & $156$ & $76.28$\\
\bottomrule 
\end{tabular}
\end{subtable}
}
\caption{Mixed precision quantization results of ResNet-$18$ and ResNet-$50$. ``Int-Only" means only including integer during quantization process. ``Uniform" represents uniform quantization. W/A is the bit-width of weight and activation. * indicates mixed precision. $\triangledown$ represents not quantizing the first and last layers.}
\label{tab:QAT resnet}
\end{table*}

\section{Experiments} \label{section:Experiments}

In this section, we conduct a series of experiments to validate the effectiveness of OMPQ on ImageNet. We first introduce the implementation details of our experiments. Ablation experiments about the monotonically decreasing function and deconstruction granularity are then conducted to demonstrate the importance of each component. Finally, we combine OMPQ with widely-used QAT and PTQ schemes, which shows a better compression and the accuracy trade-off comparing to the SOTA methods.

\subsection{Implementation Details}

The ImageNet dataset includes 1.2M training data and 50,000 validation data. 
We randomly obtain 64 training data samples for ResNet-$18$/$50$ and 32 training data samples for MobileNetV$2$ following similar data pre-processing \cite{he2016deep} to derive the set of functions $ \mathcal{F} $. 
OMPQ is extremely efficient which only needs a piece of Nvidia Geforce GTX 1080Ti and a single Intel(R) Xeon(R) CPU E5-2620 v4. For the models that have a large amount of parameters, we directly adopt the round function to convert the bit-width into an integer after linear programming. Meanwhile, 
we adopt depth-first search (DFS) to find the bit configuration that strictly meets the different constraints for a small model, \emph{e.g.}~ResNet-$18$. The aforementioned processes are extremely efficient, which only take a few seconds on these devices. Besides, OMPQ is flexible, which is capable of leveraging different search spaces with QAT and PTQ under different requirements. Finetuning implementation details are listed as follows.

For the experiments on QAT quantization scheme, we use two NVIDIA Tesla V100 GPUs. Our quantization framework does not contain integer division or floating point numbers in the network. 
In the training process, the initial learning rate is set to 1e-4, and the batch size is set to $128$. We use cosine learning rate scheduler and SGD optimizer with 1e-4 weight decay during $90$ epochs without distillation. We fix the weight and activation of first and last layer at $8$ bit following previous works, where the search space is 4-8 bit.

For the experiments on PTQ quantization scheme, we perform OMPQ on an NVIDIA Geforce GTX 1080Ti and combine it with the finetuning block reconstruction algorithm BRECQ. In particular, the activation precision of all layers are fixed to 8 bit. In other words, only the weight bit is searched, which is allocated in the 2-4 bit search space.

\subsection{Ablation Study}
\label{Ablation Study}

\noindent \textbf{Monotonically Decreasing Function.} We then investigate the monotonically decreasing function in Eq.~\ref{eq:layer_imp}. Obviously, the second-order derivatives of monotonically decreasing functions in Eq.~\ref{eq:layer_imp} influence the changing rate of orthogonality differences. In other words, the variance of the orthogonality between different layers becomes larger as the rate becomes faster. We test the accuracy of five different monotonically decreasing functions on quantization-aware training of ResNet-$18$ (6.7Mb) and post-training quantization of MobileNetV$2$ (0.9Mb). 
We fix the activation to $8$ bit.


\begin{table*}[htb]
\centering

\setlength{\tabcolsep}{5.1mm}{
\begin{subtable}[ResNet-$18$]{1\linewidth}
\caption{ResNet-$18$}
\centering
\begin{tabular}{cccccc}
\toprule  
\multirow{2}{*}{Method} & 
\multirow{2}{*}{W/A} &
\multirow{2}{*}{Model Size (Mb)}&
\multirow{2}{*}{Top-1 (\%)}&
Searching &
Searching \\
~ & ~ & ~ & ~ & Data & Iterations \\
\midrule  
Baseline & $32$/$32$ & $44.6$  & $71.08$ &-&-\\
\midrule  
FracBits-PACT & $*$/$*$ & $4.5$  & $69.10$ & $1.2$M&$120$\\
OMPQ & $*$/$4$ & $4.5$  & $68.69$ & $\textbf{64}$ & $\textbf{0}$ \\
OMPQ & $*$/$8$ & $4.5$  & $\textbf{69.94}$ & $\textbf{64}$ & $\textbf{0}$ \\
\midrule  
ZeroQ & $4$/$4$ & $5.81$  & $21.20$ &-&-\\
$\text{BRECQ}^\dag$ & $4$/$4$ & $5.81$  & $69.32$ &-&-\\
PACT & $4$/$4$ & $5.81$  & $69.20$ &-&-\\
HAWQ-V3 & $4$/$4$ & $5.81$ & $68.45$ &-&-\\
FracBits-PACT & $*$/$*$ & $5.81$  & $\textbf{69.70}$ & $1.2$M & $120$\\
OMPQ & $*$/$4$ & $5.5$ & $69.38$ & $\textbf{64}$ & $\textbf{0}$ \\
\midrule  
BRECQ & $*$/$8$ & $4.0$ & $68.82$ & $1,024$ & $100$\\
OMPQ & $*$/$8$ & $4.0$ & $\textbf{69.41}$ & $\textbf{64}$ & $\textbf{0}$ \\
\bottomrule 
\end{tabular}
\end{subtable}}
\setlength{\tabcolsep}{6.0mm}{
\begin{subtable}[MobileNetV$2$]{1\linewidth}
\centering
\caption{MobileNetV$2$}
\centering
\begin{tabular}{cccccc}
\toprule  
\multirow{2}{*}{Method} & 
\multirow{2}{*}{W/A} & 
\multirow{2}{*}{Model Size (Mb)}&
\multirow{2}{*}{Top-1 (\%)}&
Searching &
Searching \\
~ & ~ & ~ & ~ & Data & Iterations \\
\midrule  
Baseline & $32$/$32$ & $13.4$ & $72.49$ & - & - \\
\midrule  
BRECQ & $*$/$8$ & $1.3$ & $68.99$ & $1,024$ & $100$ \\
OMPQ & $*$/$8$ & $1.3$ & $\textbf{69.62}$ & $\textbf{32}$ & $\textbf{0}$ \\
\midrule  
FracBits & $*$/$*$ & $1.84$  & $69.90$ & $1.2$M & $120$ \\
BRECQ & $*$/$8$ & $1.5$ & $70.28$ & $1,024$ & $100$ \\
OMPQ & $*$/$8$ & $1.5$ & $\textbf{71.39}$ & $\textbf{32}$ & $\textbf{0}$ \\

\bottomrule 
\end{tabular}
\end{subtable}
}
\caption{Mixed precision post-training quantization experiments on ResNet-$18$ and MobileNetV$2$. $\dag$ means using distilled data in the finetuning process.}
\label{tab:PTQ}
\end{table*}

It can be observed from Table~\ref{tab:Decreasing Function} that the accuracy gradually decreases with the increasing of changing rate.
For the corresponding bit configuration, we also observe that a larger changing rate also means a more aggressive bit allocation strategy. 
In other words, OMPQ tends to assign more different bits between layers under a large changing rate, which leads to worse performance in network quantization.
Another interesting observation is the accuracy on ResNet-$18$ and MobileNetV$2$. 
Specifically, quantization-aware training on ResNet-$18$ requires numerous data, which makes the change of accuracy insignificant. 
On the contrary, post-training quantization on MobileNetV$2$ is incapable of assigning bit configuration that meets the model constraints when the functions are set to $ -x^3 $ or $ -e^x $. To this end, we select $ e^{-x} $ as our monotonically decreasing function in the following experiments.

\noindent \textbf{Deconstruction Granularity.} We study the impact of different deconstruction granularity on model accuracy. Specifically, we test four different granularity including layer-wise, block-wise, stage-wise and net-wise on the quantized-aware training of ResNet-$18$ and the post-training quantization of MobileNetV$2$. As reported in Table~\ref{tab:granularity}, the accuracy of the two models is increasing with finer granularities. Such difference is more significant on MobileNetV$2$ due to the different sensitiveness between the point-wise and depth-wise convolution. We thus employ layer-wise granularity in the following experiments.

\subsection{Quantization-Aware Training}
\label{Quantization-Aware Training}

We perform quantization-aware training on ResNet-$18$/$50$, where the results and compress ratio are compared with the previous unified quantization methods \cite{park2018value,choi2018pact,zhang2018lq} and mixed precision quantization \cite{wang2019haq,chin2020one,yao2021hawq}. 
As shown in Table~\ref{tab:QAT resnet}, OMPQ shows the best trade-off between accuracy and compress ratio on ResNet-$18$/$50$. For example, we achieve $72.08\%$ on ResNet-$18$ with only $6.7$Mb and $75$BOPs. Comparing with HAWQ-V3\cite{yao2021hawq}, the difference of the model size is negligible ($6.7$Mb, $75$BOPs vs $6.7$Mb, $72$BOPs). Meanwhile, the model compressed by OMPQ is $1.86\%$ higher than HAWQ-V3\cite{yao2021hawq}. Similarly, we achieve $76.28\%$ on ResNet-$50$ with $18.7$Mb and $156$BOPs. OMPQ is $0.89\%$ higher than HAWQ-V3 with similar model size ($18.7$Mb, $156$BOPs vs $18.7$Mb, $154$BOPs).

\begin{figure}[tb]
\centering
\includegraphics[width=1.0\linewidth]{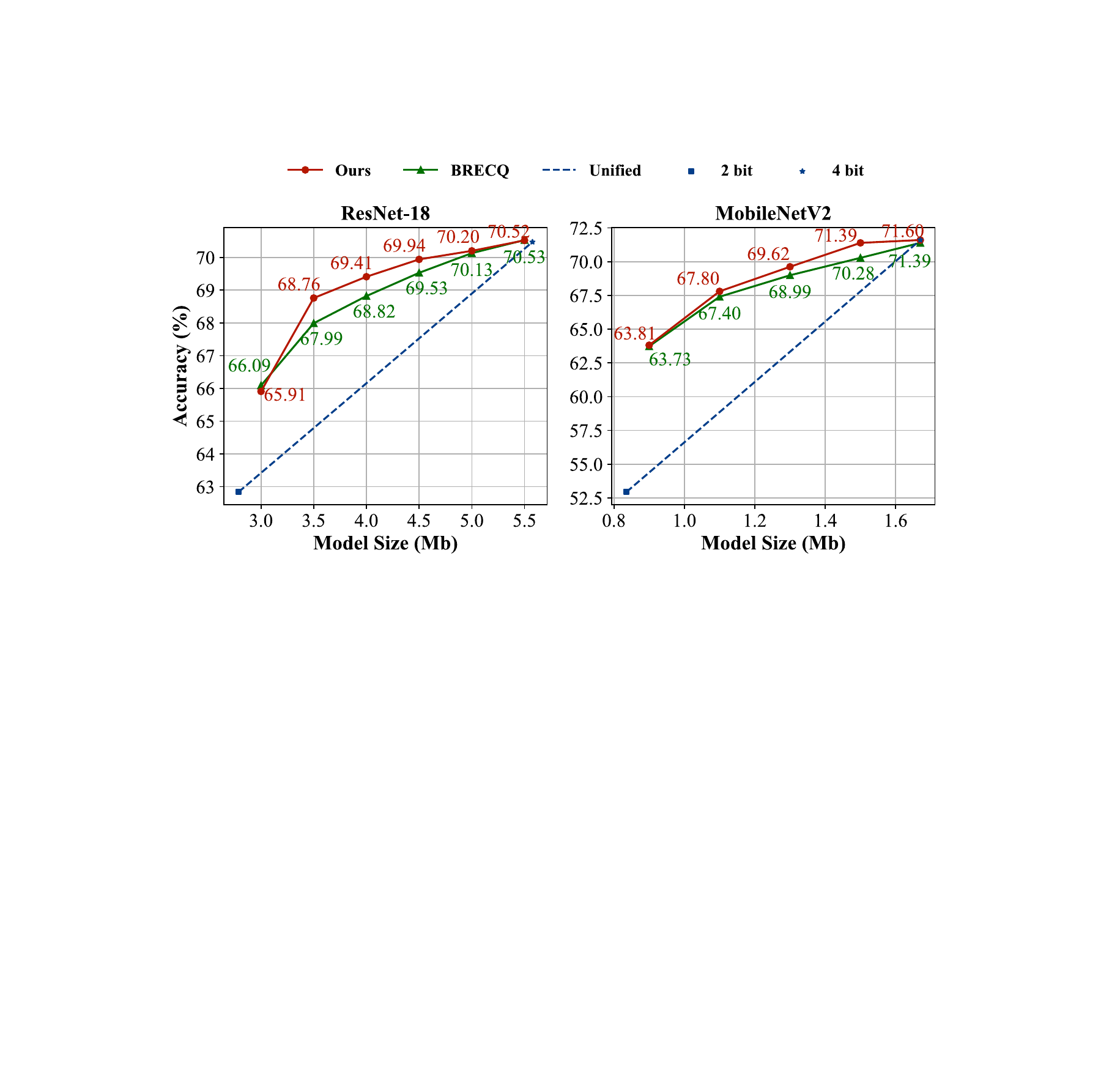}
\caption{Mixed precision quantization comparison of OMPQ and BRECQ on ResNet-$18$ and MobileNetV$2$.}
\label{fig:OMPQ vs BRECQ}

\end{figure}

\subsection{Post-Training Quantization}
\label{Post-Training Quantization}
As we mentioned before, OMPQ can also be combined with PTQ scheme to further improve the quantization efficiency thanking to its low data dependence and search efficiency. 
Previous PTQ method BRECQ \cite{li2021brecq} proposes block reconstruction quantization strategy to reduce quantization errors. 
We replace the evolutionary search algorithm with OMPQ and combine it with the finetuning process of BRECQ, which rapidly reduces the search cost and achieves better performance.
Experiment results are demonstrated in Table~\ref{tab:PTQ}, we can observe that OMPQ clearly shows the superior performance to unified quantization and mixed precision quantization methods under different model constraints. 
In particular, OMPQ outperforms BRECQ by $0.52\%$ on ResNet-$18$ under the same model size ($4.0$Mb). 
OMPQ also outperforms FracBits by $1.37\%$ on MobileNetV$2$ with a smaller model size ($1.5$Mb vs $1.8$Mb).

We also compare OMPQ with BRECQ and unified quantization,
where the results are reported in Fig.~\ref{fig:OMPQ vs BRECQ}. 
Obviously, the accuracy of OMPQ is generally higher than BRECQ on ResNet-$18$ and MobileNetV$2$ with different model constraints. Furthermore, OMPQ and BRECQ are both better than unified quantization, which shows that mixed precision quantization is superior. 


	

			

\section{Conclusion}
In this paper, we have proposed a novel mixed precision algorithm, termed OMPQ, to effectively search the optimal bit configuration on the different constraints. Firstly, we derive the orthogonality metric of neural network by generalizing the orthogonality of the function to the neural network. Secondly, we leverage the proposed orthogonality metric to design a linear programming problem, which is capable of finding the optimal bit configuration. 
Both orthogonality generation and linear programming solving are extremely efficient, which are finished within a few seconds on a single CPU and GPU. 
Meanwhile, OMPQ also outperforms the previous mixed precision quantization and unified quantization methods. 
Furthermore, we will explore the mixed precision quantization method combining multiple knapsack problem with the network orthogonality metric.

\section*{Acknowledgments}

This work is supported by the National Science Fund for Distinguished Young (No.62025603), the National Natural Science Foundation of China (No.62025603, No. U1705262, No. 62072386, No. 62072387, No. 62072389, No. 62002305,  No.61772443, No. 61802324 and No. 61702136), Guangdong Basic and Applied Basic Research Foundation (No.2019B1515120049), China National Postdoctoral Program for Innovative Talents (BX20220392), and China Postdoctoral Science Foundation (2022M711729).

\bibliography{aaai23}

\clearpage

\section*{Supplementary Material}

\section{Calculation Acceleration}

Suppose that $f_i(X)\in\mathbb{R}^{N\times(C_i\times H_i\times W_i)}$ and $f_j(X)\in\mathbb{R}^{N\times(C_j\times H_j\times W_j)}$ are the output of the $i$-th layer and the $j$-th layer in the neural network, respectively. We set $Y=f_i(X)$, $Z=f_j(X)$, then $YY^T$, $ZZ^T\in\mathbb{R}^{N\times N}$. The calculation is accelerated through:
\begin{equation}\label{eq:efficiency_equal1}
	||Z^TY||_F^2 = \left\langle \textbf{vec}(YY^T),\textbf{vec}(ZZ^T) \right\rangle.
\end{equation}

\begin{proof}
We set $p_1=C_i\times H_i\times W_i , p_2=C_j\times H_j\times W_j$, and $z_{i,m}, y_{i,m}$ are the $i$-th row and $m$-th column entry of the matrix $Z$ and $Y$, then
\begin{equation}
\begin{split}
    & ||Z^TY||_F^2 \nonumber \\
    & =\sum_{i=1}^{p_2}\sum_{j=1}^{p_1}{\left( \sum_{m=1}^N z_{m,i}y_{m,j} \right)}^2 \\
    & =\sum_{i=1}^{p_2}\sum_{j=1}^{p_1}\left( \sum_{m=1}^N z_{m,i}y_{m,j} \right)\left( \sum_{n=1}^N z_{n,i}y_{n,j} \right) \\
    & =\sum_{i=1}^{p_2}\sum_{j=1}^{p_1}\sum_{m=1}^N\sum_{n=1}^N z_{m,i}y_{m,j}z_{n,i}y_{n,j} \\
    & =\sum_{m=1}^N\sum_{n=1}^N\sum_{i=1}^{p_2}\sum_{j=1}^{p_1} z_{m,i}y_{m,j}z_{n,i}y_{n,j} \\
    & =\sum_{m=1}^N\sum_{n=1}^N \left(\sum_{j=1}^{p_1}y_{m,j}y_{n,j}\right)\left(\sum_{i=1}^{p_2} z_{m,i}z_{n,i}\right) \\
    &= \left\langle \textbf{vec}(YY^T),\textbf{vec}(ZZ^T) \right\rangle.
\end{split}
\end{equation}

\end{proof}
\noindent From Eq.~\ref{eq:efficiency_equal1}, we can see that two computation forms have different time complexities when dealing with different situations. 
Specifically, the time complexities of calculating $||Z^TY||_F^2$ and inner product of vectors are $\mathbf{O}(Np_1p_2)$ and $\mathbf{O}(N^2(p_1+p_2+1))$, respectively. 
In other words, when feature number $(p_1~\text{or}~p_2)$ is larger than the number of samples $N$, we take the inner product form to speed up the calculation. 
Conversely, using $||Z^TY||_F^2$ is clearly more faster than inner product form. 

Concretely, we take $N,p \in \{10000,100\}$ as an example and randomly generate matrix $Y,Z \in \mathbb{R}^{N\times p}$. 
In our experiment, we calculate the ORM of $Y, Z$ in vector inner product form and norm form, respectively. 
The results are reported in Table~\ref{tab:calculation time}. 
From this table, when the number of features is less than the number of samples, calculating ORM in the norm form is much faster (54$ \times $) than the inner product form. 
On the contrary, when the number of features is greater than the number of samples, the inner product form calculates the ORM faster (70$ \times $).

\begin{table}[htb]
\begin{center}

\setlength{\tabcolsep}{2.1mm}{
\begin{tabular}{cccc}
\toprule  
 & \multicolumn{2}{c}{\textbf{Calculation Strategy}} & \multirow{2}{*}{\textbf{Acceleration Ratio}} \\
\cline{2-3}
& \textbf{Inner Product} & \textbf{Norm} & \\
\midrule  
\textbf{$N$\textgreater $p$} & $18.42$& $0.34$& $54\times$\\
\textbf{$N$\textless $p$}& $0.12$& $8.40$&$70\times$\\
\bottomrule 
\end{tabular}}
\caption{The ORM calculation time (second) of different calculation strategy between matrix $Y$ and $Z$ which have the same size. There are two cases for the size of $Y$ or $Z$, \emph{i.e.}, feature number $ p $ is larger/smaller than the number of the samples $ N $.}
\label{tab:calculation time}
\end{center}
\end{table}







\section{Network Deconstruction and Linear Programming}
In this section, we elaborate on why the network is deconstructed by the former $i$ layers instead of the $i$-th layer. 
Furthermore, we will illustrate how to  construct a linear programming problem in details.
\subsection{Network Deconstruction}
\begin{figure}[tb]
	\centering
	\includegraphics[width=1.0\linewidth]{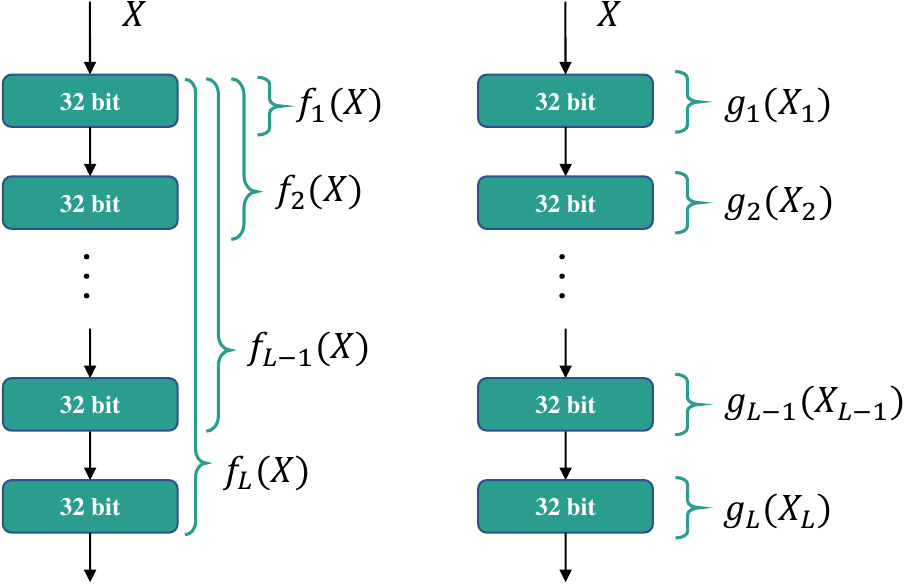}
	\caption{The former $i$ layers deconstruction(left) and the $i$-th layer deconstruction(right). We omit the BatchNorm layer \cite{ioffe2015batch} and the activation layer for simplicity while $X_1=X, X_2=g_1(X_1), \dots, X_L=g_{L-1}(X_{L-1})$.}
	\label{fig:deconstruction}
\end{figure}

Fig.~\ref{fig:deconstruction} demonstrates two approaches to the deconstruction of the network: the former $i$ layers deconstruction and the $i$-th layer deconstruction, and we naturally accept that the latter deconstruction is more intuitive. We give the definition of orthogonality for the $i$-th and $j$-th layers under the latter deconstruction:
\begin{equation}
\begin{split}
	{\left\langle g_i,g_j \right\rangle}_{P(x)}
	&= \int_{\mathcal{D}} g_i(x_i)P(x){g_j(x_j)}^T dx
	\label{eq:montecarlo_decon},
\end{split}
\end{equation}
where $x_i$, $x_j$ are the outputs of the former layer, respectively. To unify the inputs, we further expand Eq.~\ref{eq:montecarlo_decon} as follows:
\begin{equation}
\begin{split}
	&{\left\langle g_i,g_j \right\rangle}_{P(x)}\\
	&= \int_{\mathcal{D}} g_i(g_{i-1}(\dots g_1(x)))P(x){g_j(g_{j-1}(\dots g_1(x)))}^T dx
	\label{eq:unify_output}.
\end{split}
\end{equation}
To facilitate the mathematical analysis, we set $f_i(x) = g_i(g_{i-1}(\dots g_1(x)))$, $f_j(x) = g_j(g_{j-1}(\dots g_1(x)))$ and the composition function $f(x)$ represents the former $i$ layers of the network. Therefore, the unified input restricts us to deconstruct the network according to the former $i$ layers only. Moreover, two deconstruction approaches are equivalent in the context of unified input. Having determined the network deconstruction, we next explain that how to construct a  linear programming problem associate to the network deconstruction.

\begin{figure}[tb]
	\centering
	\includegraphics[width=1.0\linewidth]{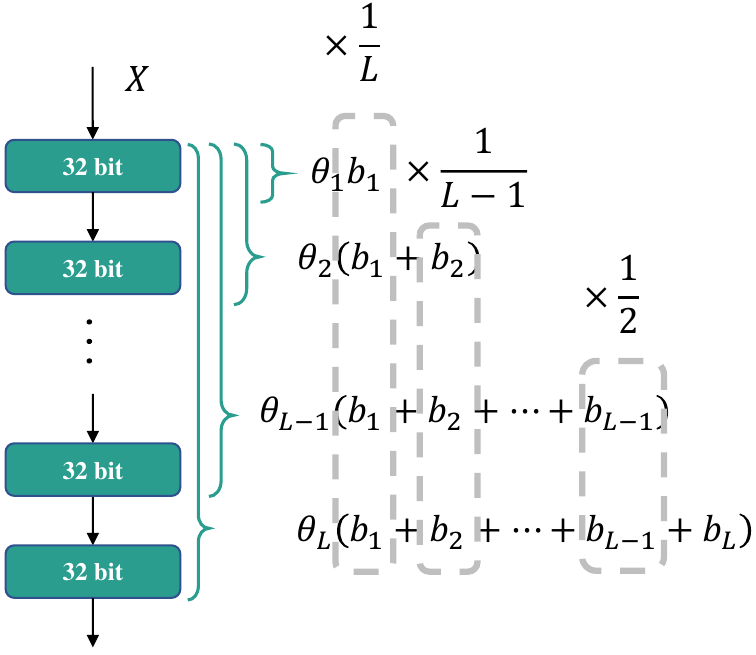}
	\caption{Compositions of linear programming objective function. $b_i$ represents the bit-width variable of the $i$-th layer. $\theta_i$ represents the importance factor of the former $i$ layers.}
	\label{fig:linear_programming}
\end{figure}

\begin{figure*}[t]
	\centering
	\includegraphics[width=1.0\linewidth]{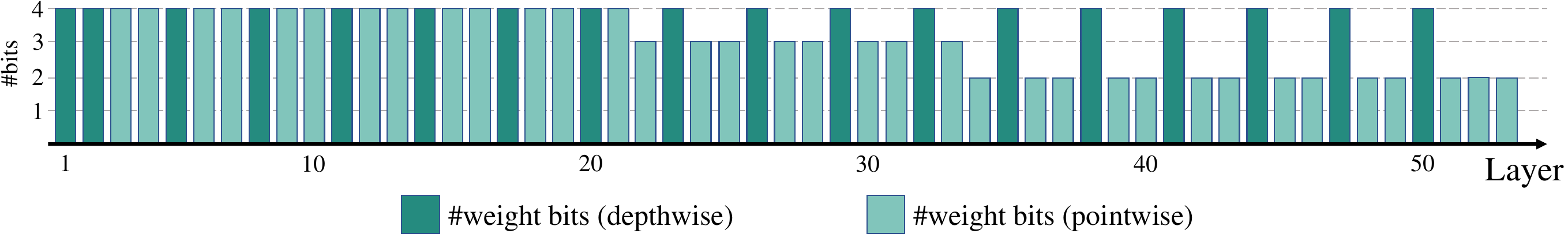}
	\caption{Layerwise bit configuration of MobileNetV$2$, which is quantized to $0.9$Mb}
	\label{fig:mbv2_bit_config}
\end{figure*}

\begin{figure}[tb]
	\centering
	\includegraphics[width=0.8\linewidth]{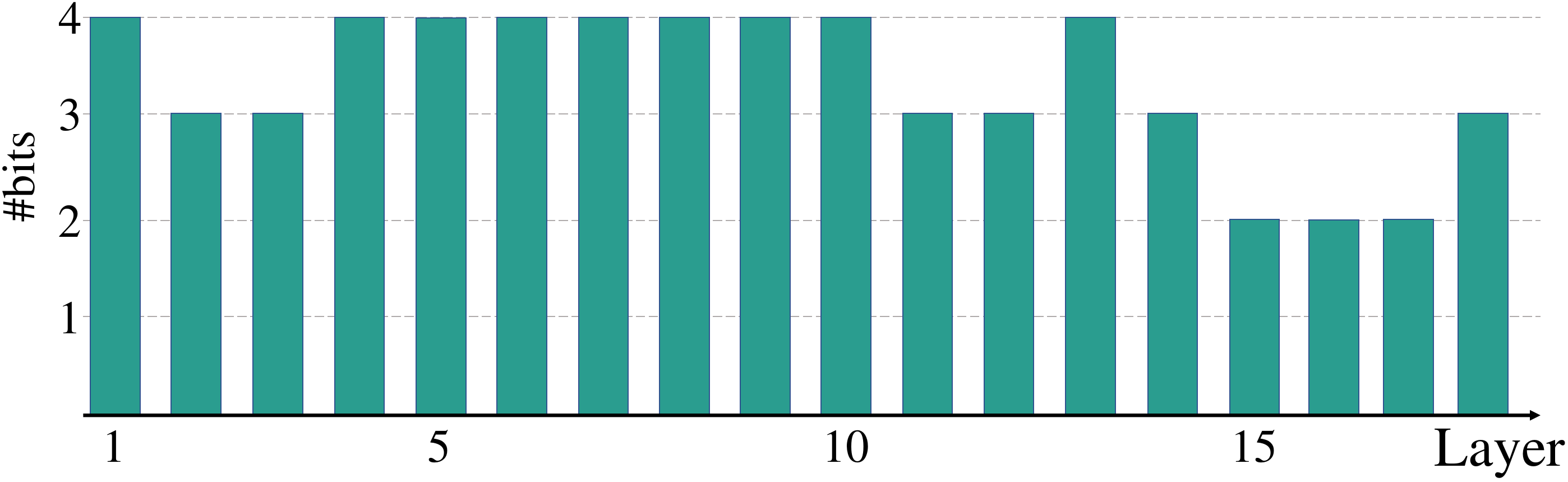}
	\caption{Layerwise bit configuration of ResNet-$18$, which is quantized to $3.5$Mb}
	\label{fig:res_bit_config}
\end{figure}

\begin{figure}[tb]
	\centering
	\includegraphics[width=1.0\linewidth]{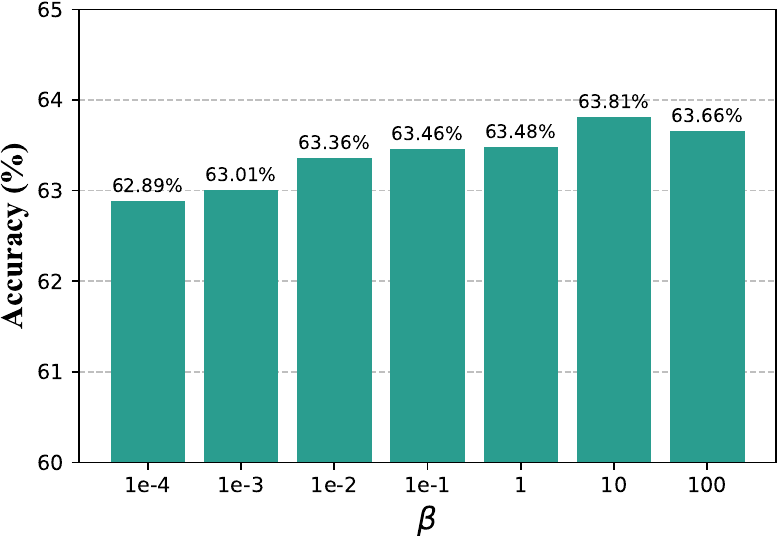}
	\caption{The relationship between accuracy and hyperparameter $\beta$ on MobileNetV$2$, which is quantized to $0.9$Mb.}
	\label{fig:beta}
\end{figure}

\subsection{Linear Programming}
\label{Sec:Linear Programming}
According to Eq.~\ref{eq:layer_imp1}, we calculate the importance factor $\theta_i$ corresponding to the former $i$ layers of the network,
\begin{equation}\label{eq:layer_imp1}
	\theta_i = e^{-\beta\gamma_i}.
\end{equation}
The stronger orthogonality of the former $i$ layers implies that the importance factor $\theta_i$ is larger. Therefore, we take $\theta_i$ as the weight coefficient for the bit-width assignment of the former $i$ layers, which means that the important layers are assigned a larger bit-width to retain more information. However, this will lead to different scales of weight coefficient for different layers due to accumulation, \emph{e.g.}~the range of weight coefficients for the first layer and the last layer is $[Le^{-\beta(L-1)},L]$ and $[e^{-\beta(L-1)},1]$, respectively. We thus rescale the $i$-th layer by multiplying the factor $1/(L-i+1)$. Finally, we sum up all the terms to obtain the objective function of the linear programming in Eq.~\ref{linear_pro1}. More details are intuitively illustrated in Fig.~\ref{fig:linear_programming}. 

\begin{equation}
\begin{split}
\label{linear_pro1}
	\text{Objective:    }
	     &\max_{\mathbf{b}} \sum_{i=1}^{L}\left( \frac{b_i}{L-i+1} \sum_{j=i}^{L} \theta_j \right), \\
	\text{Constraints:    }
	     &  \sum_i^L M^{(b_i)} \leq \mathcal{T}.
\end{split}
\end{equation}

\section{ORM Matrix}
\begin{figure*}[t]
	\centering
	\includegraphics[width=0.85\linewidth]{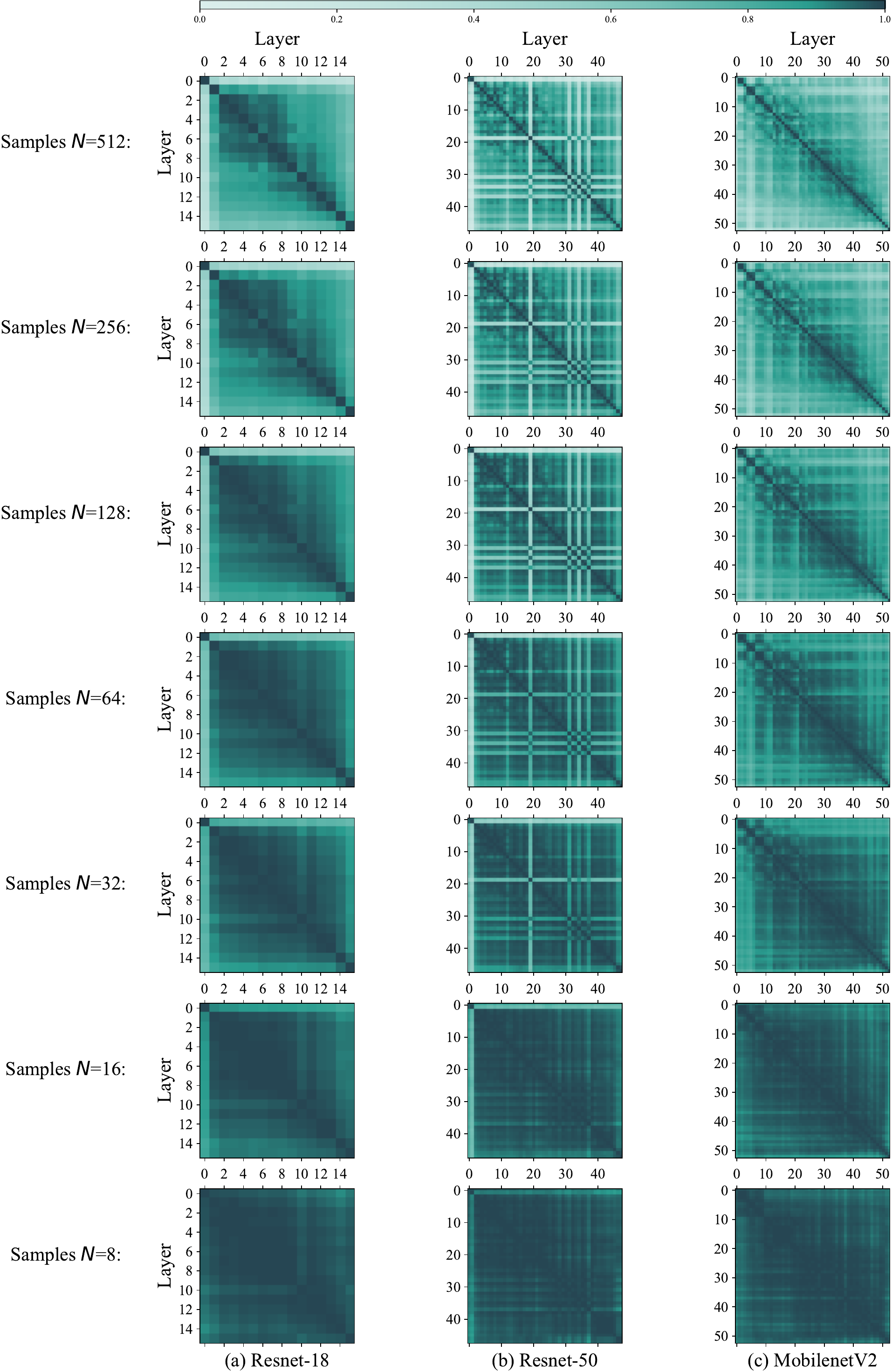}
	\caption{The ORM matrices of ResNet-$18$, ResNet-$50$ and MobileNetV$2$ with different samples $N$.}
	\label{fig:ORM_matrix}
\end{figure*}

We present the ORM matrices of ResNet-$18$, ResNet-$50$ and MobileNetV$2$ for different samples $N$ in Fig.~\ref{fig:ORM_matrix}. We can obtain the following conclusions: (a) the orthogonality between the former $i$ and the former $j$ layers generally becomes stronger as the absolute difference $|i-j|$ increases. (b) the orthogonality discrepancy of the network which output activation lies within the same block is tiny. (c) as the samples $N$ increases, the approximation of expectation becomes more accurate and the orthogonality discrepancy of different layers becomes more and more obvious.

\section{Importance Variance}
According to Sec~\ref{Sec:Linear Programming}, the hyperparameter $\beta$ may also affect the range of the weight coefficients of bit-width variable as follows:
\begin{equation}
\begin{split}
\label{range1}
	\left[\lim_{\beta \to +\infty}e^{-\beta(L-1)},1\right]&=\left[0,1\right],\\
	\left[\lim_{\beta \to 0^+}e^{-\beta(L-1)},1\right]&=\left\{1\right\}.
\end{split}
\end{equation}

From Eq.~\ref{range1}, when $\beta$ increases, the range of importance factor $\theta$ becomes larger. On the contrary, when $\beta$ approaches zero, the range of $\theta$ is greatly compressed, so the variance of importance factors of different layers is greatly reduced. We explore the relationship between accuracy and $\beta$ on MobileNetV$2$. We quantize MobileNetV$2$ to $0.9$Mb and fix all activation value at $8$ bit. As shown in Fig.~\ref{fig:beta}, when $\beta$ increases, the accuracy gradually increases as the variance of importance factors of different layers becomes larger, and stabilizes when $\beta$ is large enough. According to our experiments, the low variance of importance factor $\theta$ may lead the linear programming algorithm to choose an aggressive bit configuration resulting in sub-optimal accuracy.

\begin{figure}[tb]
\centering
\includegraphics[width=1.0\linewidth]{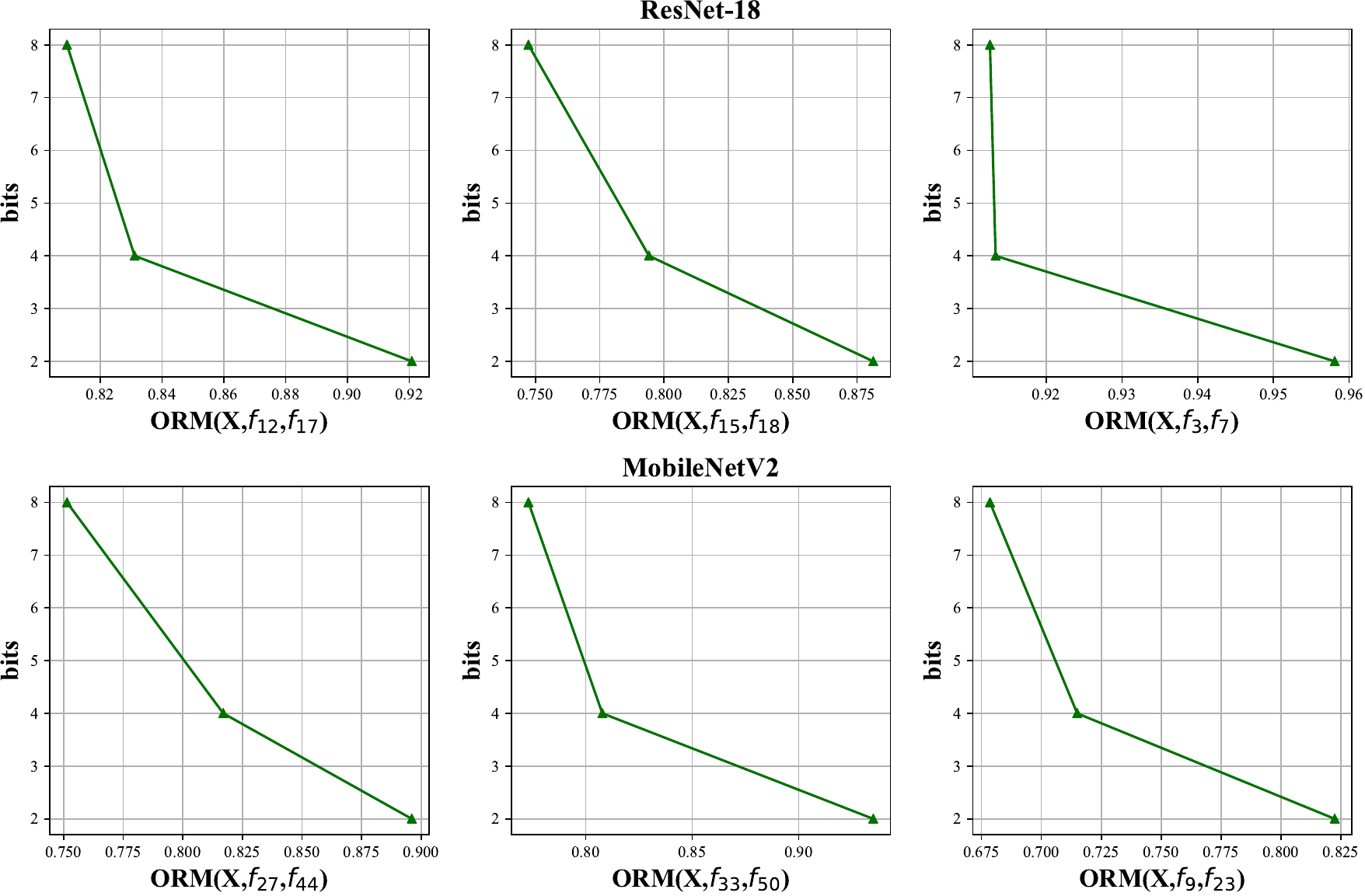}
\caption{Relationship between orthogonality and bit-width of different layers on ResNet-$18$ and MobileNetV$2$.}
\label{fig:orthogonality_bit-width}
\end{figure}

\section{Bit Configurations}

We also give the layer-wise bit configurations of ResNet-$18$ \cite{he2016deep} and MobileNetV$2$ \cite{howard2017mobilenets} on PTQ. As shown in Fig.~\ref{fig:mbv2_bit_config}-\ref{fig:res_bit_config}, we observe that MobileNetV$2$ allocates more bits to the depthwise convolution, and allocates fewer bits to the pointwise convolution. In other words, the $3\times3$ convolutional layers are more orthogonal than the $1\times1$ convolutional layers, which somehow explains the fact that networks containing a large number of $1\times1$ convolutional layers are difficult to quantize. Therefore, this phenomenon can potentially be instructive for low precision quantization and binarization of ResNet-$50$, MobileNetV$2$ and even Transformer \cite{dosovitskiy2021an}. Besides, the bit-width of each layer on ResNet-$18$ and MobileNetV$2$ decreases as the depth of the layer increases.

\section{Effectiveness of ORM on Mixed Precision Quantization}

We have already provided the positive correlation between model orthogonality and quantization accuracy in the paper. We will further reveal the relationship between layer orthogonality and bit-width in Fig.~\ref{fig:orthogonality_bit-width}. We randomly select some layers and compute their orthogonality under different bit-widths. Obviously, layer orthogonality is positively correlated with bit-width on different models. Furthermore, The positive correlation coefficient of orthogonality and bit width is different for different layers, which implies that the sensitivity of orthogonality is different for different layers. Therefore, the positive correlation and sensitivity differences between orthogonality and bit-width allow us to construct the objective function for the linear programming problem.

\section{Derivation of ORM}

Leveraging Monte Carlo sampling, we obtain the following approximation

\begin{equation}\label{eq:Intermediate_appendix}
	N\int_{\mathcal{D}} f_i(x)P(x){f_j(x)}^T dx \approx {\left\Vert f_j(X)^{T}f_i(X)\right\Vert}_F. 
\end{equation}

\noindent Applying Cauchy-Schwarz inequality to the left side of Eq.~\ref{eq:Intermediate_appendix}, we have

\begin{equation}
\begin{split}
\label{eq:Intermediate1_appendix}
	0 &\leq {\left( N\int_{\mathcal{D}} f_i(x)P(x){f_j(x)}^T dx \right)}^2 \\
	& \leq \int_{\mathcal{D}} Nf_i(x)P_i(x){f_i(x)}^T dx \int_{\mathcal{D}} Nf_j(x)P_j(x){f_j(x)}^T dx. 
\end{split}
\end{equation}

\noindent Normalize Eq.~\ref{eq:Intermediate1_appendix} to $[0,1]$, we have 

\begin{equation}
\begin{split}
\begin{aligned}
\label{eq:Intermediate2_appendix}
	&\frac{{\left( N\int_{\mathcal{D}} f_i(x)P(x){f_j(x)}^T dx \right)}^2}{\int_{\mathcal{D}} Nf_i(x)P_i(x){f_i(x)}^T dx \int_{\mathcal{D}} Nf_j(x)P_j(x){f_j(x)}^T dx} \\
	&\in [0, 1]. 
\end{aligned}
\end{split}
\end{equation}

\noindent We substitute Eq.~\ref{eq:Intermediate_appendix} into Eq.~\ref{eq:Intermediate2_appendix} and obtain ORM

\begin{equation}\label{eq:ORM1}
\begin{aligned}
	&{\rm ORM}(X,f_i,f_j) \\
	&=  \frac{{||f_j(X)^{T}f_i(X)||}^2_F}{||f_i(X)^{T}f_i(X)||_F||f_j(X)^{T}f_j(X)||_F} \in [0,1]. 
\end{aligned}
\end{equation}

\end{document}